\documentclass[conference]{IEEEtran}
\IEEEoverridecommandlockouts
\usepackage{cite}
\usepackage{amsmath,amssymb,amsfonts}
\usepackage{algorithmic}
\usepackage{graphicx}
\usepackage{textcomp}
\usepackage{xcolor}

\def\BibTeX{{\rm B\kern-.05em{\sc i\kern-.025em b}\kern-.08em
    T\kern-.1667em\lower.7ex\hbox{E}\kern-.125emX}}

\usepackage{subfig}

\begin{document}

\title{Improving the Spatial Resolution of GONG Solar Images to GST Quality Using Deep Learning\\
\thanks{*Dr. Bo Shen is the corresponding author.}
}

\author{
\IEEEauthorblockN{Chenyang Li}
\IEEEauthorblockA{\textit{Department of Mechanical} \\
\textit{and Industrial Engineering} \\
\textit{NJIT}, Newark, USA\\
cl237@njit.edu}
\and
\IEEEauthorblockN{Qin Li}
\IEEEauthorblockA{\textit{Department of Physics} \\
\textit{NJIT}, Newark, USA\\
ql47@njit.edu}
\and
\IEEEauthorblockN{Haimin Wang}
\IEEEauthorblockA{\textit{Department of Physics} \\
\textit{NJIT}, Newark, USA\\
haimin.wang@njit.edu}
\and
\IEEEauthorblockN{Bo Shen*}
\IEEEauthorblockA{\textit{Department of Mechanical} \\
\textit{and Industrial Engineering} \\
\textit{NJIT}, Newark, USA\\
bo.shen@njit.edu}
}

\maketitle 

\begin{abstract}
High-resolution (HR) solar imaging is crucial for capturing fine-scale dynamic features such as filaments and fibrils. However, the spatial resolution of the full-disk H$\alpha$ images is limited and insufficient to resolve these small-scale structures. To address this, we propose a GAN-based super-resolution approach to enhance low-resolution (LR) full-disk H$\alpha$ images from the Global Oscillation Network Group (GONG) to a quality comparable with HR observations from the Big Bear Solar Observatory/Goode Solar Telescope (BBSO/GST). We employ Real-ESRGAN with Residual-in-Residual Dense Blocks and a relativistic discriminator. We carefully aligned GONG–GST pairs. The model effectively recovers fine details within sunspot penumbrae and resolves fine details in filaments and fibrils, achieving an average mean squared error (MSE) of 467.15, root mean squared error (RMSE) of 21.59, and cross-correlation (CC) of 0.7794. Slight misalignments between image pairs limit quantitative performance, which we plan to address in future work alongside dataset expansion to further improve reconstruction quality.
\end{abstract}


\section{Introduction}
HR solar observation is the cornerstone of understanding fine-scale dynamic structures on the Sun, which play a significant role in driving space weather events that can disturb Earth’s magnetosphere and disrupt satellite communications, power grids, and navigation systems~\cite{pesnell2012solar,priest2014magnetohydrodynamics}. The HR solar telescope, the GST at BBSO, has a spatial resolution of approximately 0.029 arcseconds per pixel. It captures exceptional detail but is restricted to a limited time due to many conditions like day-night cycle, weather, atmospheric turbulence, and maintenance~\cite {cao2010scientific,li2017high}. These HR solar images enable detailed analyses of fine-scale chromospheric dynamics, which allow researchers to closely examine the sun's structures and evolution~\cite{goode2010nst}. On the other hand, full-disk solar images captured by the GONG in the H$\alpha$ spectral line offer full-day coverage of the sun, thanks to its globally distributed network of ground-based observatories. However, with a spatial resolution of approximately 1.0 arcsecond per pixel, their practical use for detailed feature analysis is limited~\cite{harvey1996global}. In this paper, we aim to consistently generate HR H$\alpha$ images by enhancing the spatial resolution of GONG full-disk observations to a comparable quality of GST. 

Image super-resolution is a powerful image processing technique that generates HR images from LR inputs. It offers a promising solution to this challenge. Traditional super-resolution methods like interpolation typically produce unsatisfactory results, which often lose critical fine details~\cite{dong2015image}. Recently, advancements in deep learning, especially convolutional neural networks (CNNs), have dramatically improved super-resolution capabilities in natural image processing, achieving remarkable clarity and detail enhancement~\cite{dong2015image}. However, previous research on solar image super-resolution has primarily focused on enhancing images obtained from the same instrument or under similar observational conditions, rather than improving resolution across significantly different observational platforms. Recent studies have successfully recovered the GST HR magnetograms from the Helioseismic and Magnetic Imager (HMI)~\cite{song2024improving,xu2025improving}, but they can not effectively capture the temporal dynamics of fine-scale solar structures. 


To the authors' knowledge, this work is the first attempt to use a deep learning approach to enhance lower-resolution H$\alpha$ images---from GONG to GST-quality resolution. This is particularly challenging given the highly dynamic nature of the chromosphere, where temporal reconstruction remains an open problem we will explore in this study. To achieve this, we first carefully cropped the LR GONG H$\alpha$ images and aligned them with the corresponding GST observations. Then, we adopt a deep learning super-resolution method called Real-ESRGAN~\cite{wang2021real} as the backbone of our super-resolution framework. Real-ESRGAN is well-suited for handling real-world image degradation. 

\section{Dataset Description}
This study leverages two distinct solar image datasets collected on 2023/08/31: LR full-disk images from the GONG and corresponding HR images obtained from the GST at BBSO. Because the temporal cadence differs between GONG and GST, we selected the closest available GONG observation to match each GST frame.

\subsection{LR Dataset (GONG H$\alpha$ Images)}
The LR data used in this study are full-disk solar images from the GONG, specifically from the BBSO site, which was selected for coordinated observation time with the GST data. These observations were taken in the H$\alpha$ central line (6562.808 Å) and have a spatial resolution of approximately 1.0 arcsecond per pixel. The images are stored in FITS format with a size of $2048 \times 2048$ pixels.

To prepare them for super-resolution training, each full-disk image was cropped to match the field of view of the corresponding HR observation from the GST. This cropping ensures spatial alignment between the LR and HR image pairs. All GONG images used in this study were acquired on 2023/08/31, between 16:35 and 01:22 UTC, covering nearly nine hours of solar activity.

\subsection{HR Dataset (BBSO GST Images)}
The HR data come from the GST at BBSO, using the Visible Imaging Spectrometer in the H$\alpha$ band. These images have a spatial resolution of approximately 0.029 arcseconds per pixel, making them well-suited for resolving fine-scale solar features such as sunspot structure and chromospheric filaments.

The GST observations were conducted during the same day, from 16:35 to 22:35 UTC, and serve as the ground truth for training and evaluating the super-resolution model. Each HR image was carefully paired with a cropped GONG image taken at nearly the same time, allowing for a one-to-one correspondence suitable for supervised learning.

\subsection{Dataset Splits and Usage}
The collected dataset from 2023/08/31 is divided into three subsets. The training set contains 281 HR-LR image pairs from 16:35 to 21:35 UTC (5 hours) used to train the Real-ESRGAN super-resolution model. The testing set consists of 63 HR-LR image pairs from 21:35 to 22:35 UTC (1 hour). Additionally, an extended evaluation set comprising 143 LR images (without corresponding HR ground truth) collected between 22:36 and 01:22 UTC (approximately 3 hours) is included to qualitatively assess the model's performance in realistic application scenarios.


\subsection{Data Alignment}

The data alignment process consists of two main stages: temporal self-alignment of the GONG LR images and spatial co-alignment between the GONG LR and GST HR images. The raw GONG full-disk images cannot be directly used due to misalignment caused by telescope jitter. While relatively minor at GONG's native resolution, these shifts become significant when mapping to the counterparts with approximately ten times finer spatial resolution. We first stabilize the GONG sequence over time, followed by spatial alignment to match the GST field of view.

\subsubsection{Initial Cropping and Region Selection}

To construct paired data, we cropped each GONG full-disk image to match the approximate field of view of the corresponding GST observation. Cropping boundaries were determined using header metadata from the GST FITS files. Due to minor inaccuracies in telescope pointing, we initially selected a slightly larger region to ensure coverage.

\begin{figure}[!htbp] \vspace{-0.10in}
  \centering
\includegraphics[width=0.45\textwidth]{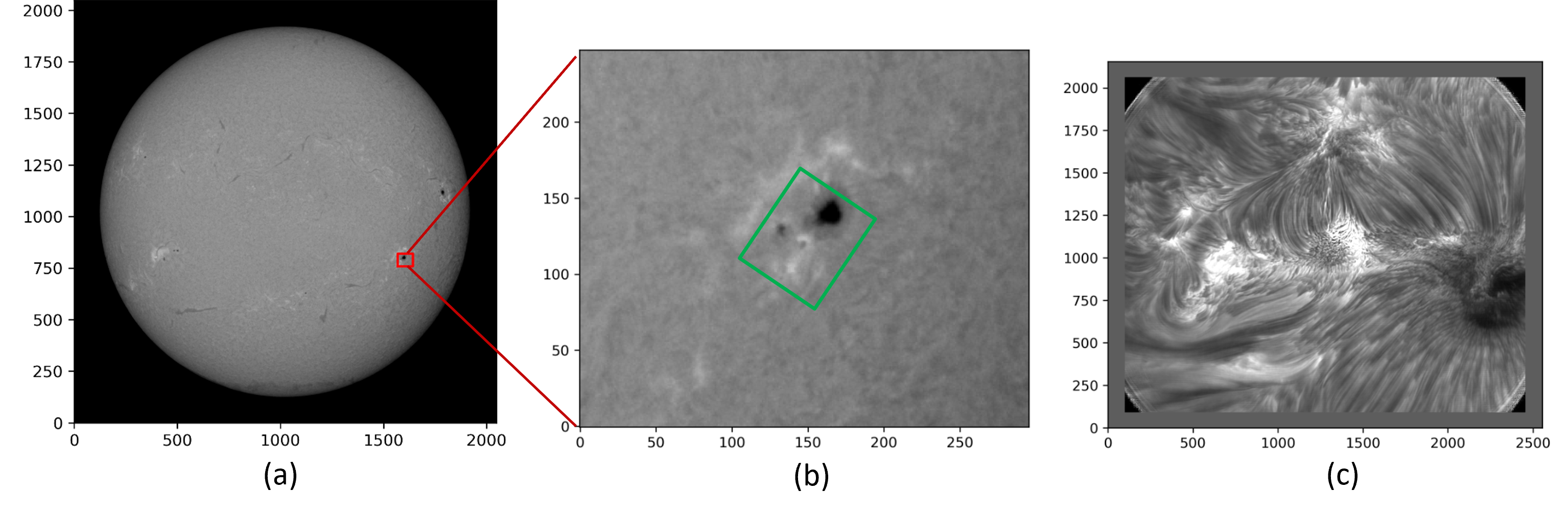} \vspace{-0.in}
  \caption{Original dataset examples: (a) Full-disk GONG image with GST coverage marked in red; (b) Cropped GONG region corresponding to GST; (c) Original GST HR observation.} \vspace{-0.15in}
  \label{fig: LR & HR original}
\end{figure}

Figure~\ref{fig: LR & HR original}a shows an example of a full-disk GONG LR image. The red box highlights the approximate large region where the corresponding GST HR is located. Figure~\ref{fig: LR & HR original}b shows the zoomed-in cropped region from the GONG image. The green box highlights the area of the corresponding GST HR observation.  Figure~\ref{fig: LR & HR original}c shows the original corresponding GST HR observation.

\subsubsection{Temporal Alignment of GONG Image Sequence}

Due to telescope jitter, the GONG images required temporal alignment. Each image was aligned to its predecessor by searching within a $\pm$30 pixel range in both x and y directions and selecting the shift maximizing the cross-correlation. The alignment process was repeated iteratively for enhanced stability.

\subsubsection{Geometric Co-Alignment with GST Observations}

Following temporal alignment, the GONG images were co-aligned with the GST images. GST images were rotated due to solar tracking; the rotation angles, recorded in the GST FITS headers, were used to rotate the GST images back into alignment with GONG. We then removed non-physical edge artifacts introduced during rotation. For precise spatial alignment, we applied the SIFT (Scale-Invariant Feature Transform) algorithm and performed minor manual adjustments.

\begin{figure}[!htbp] 
\vspace{-0.3in}
\centering
	\subfloat[Preprocessed GONG image]{\includegraphics[width=0.212\textwidth]{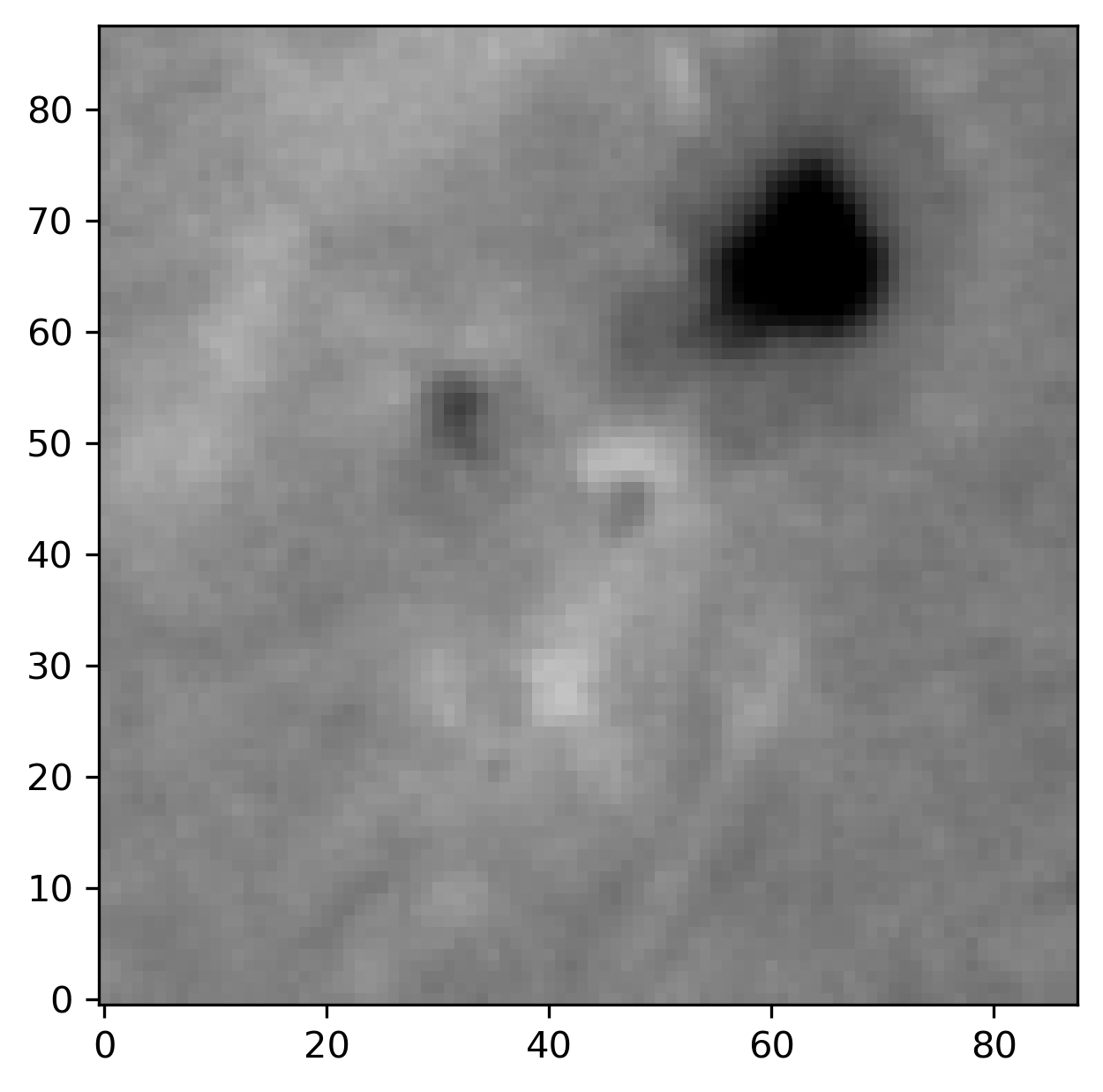} \label{subfig: GONG LR Cropped}}
    \subfloat[GST HR observation image after rotation]{\includegraphics[width=0.23\textwidth]{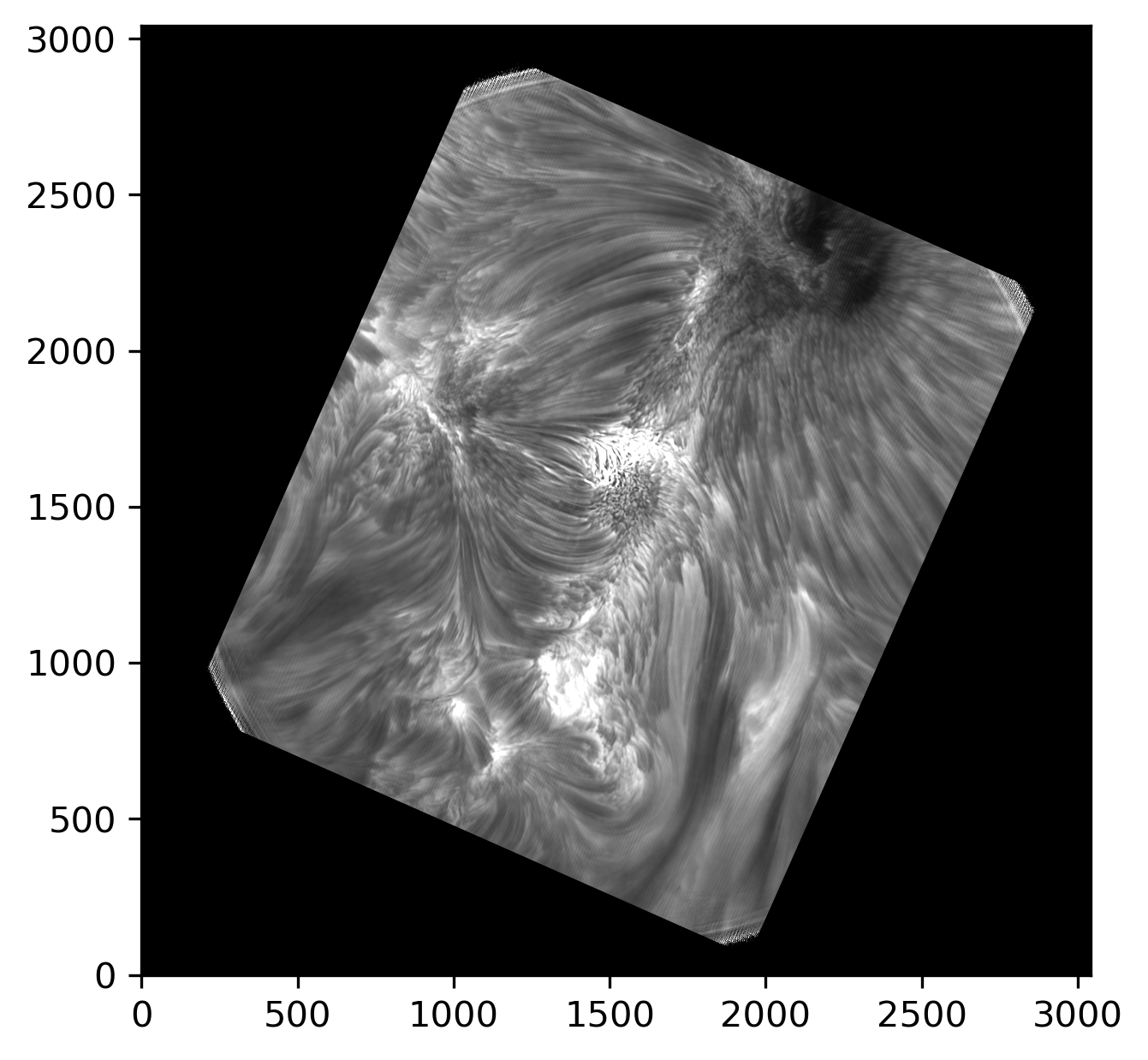} \label{subfig: GST HR Rotated}}
  \vspace{-0.05in}
\caption{Initially aligned data examples: (a) Cropped and aligned GONG LR image; (b) GST HR image after rotation adjustment and preprocessing.}  
\label{fig: LR & HR inter}
\vspace{-0.1in}
\end{figure}

Figure~\ref{subfig: GONG LR Cropped} shows a GONG LR image after cropping and temporal alignment, and Figure~\ref{subfig: GST HR Rotated} shows the GST HR image after correcting the rotation and removing edge artifacts.

\subsubsection{Final Cropping and Pair Generation}

\begin{figure}[!htbp] \vspace{-0.12in}
\centering
	\subfloat[Final cropped GONG LR image]{\includegraphics[width=0.22\textwidth]{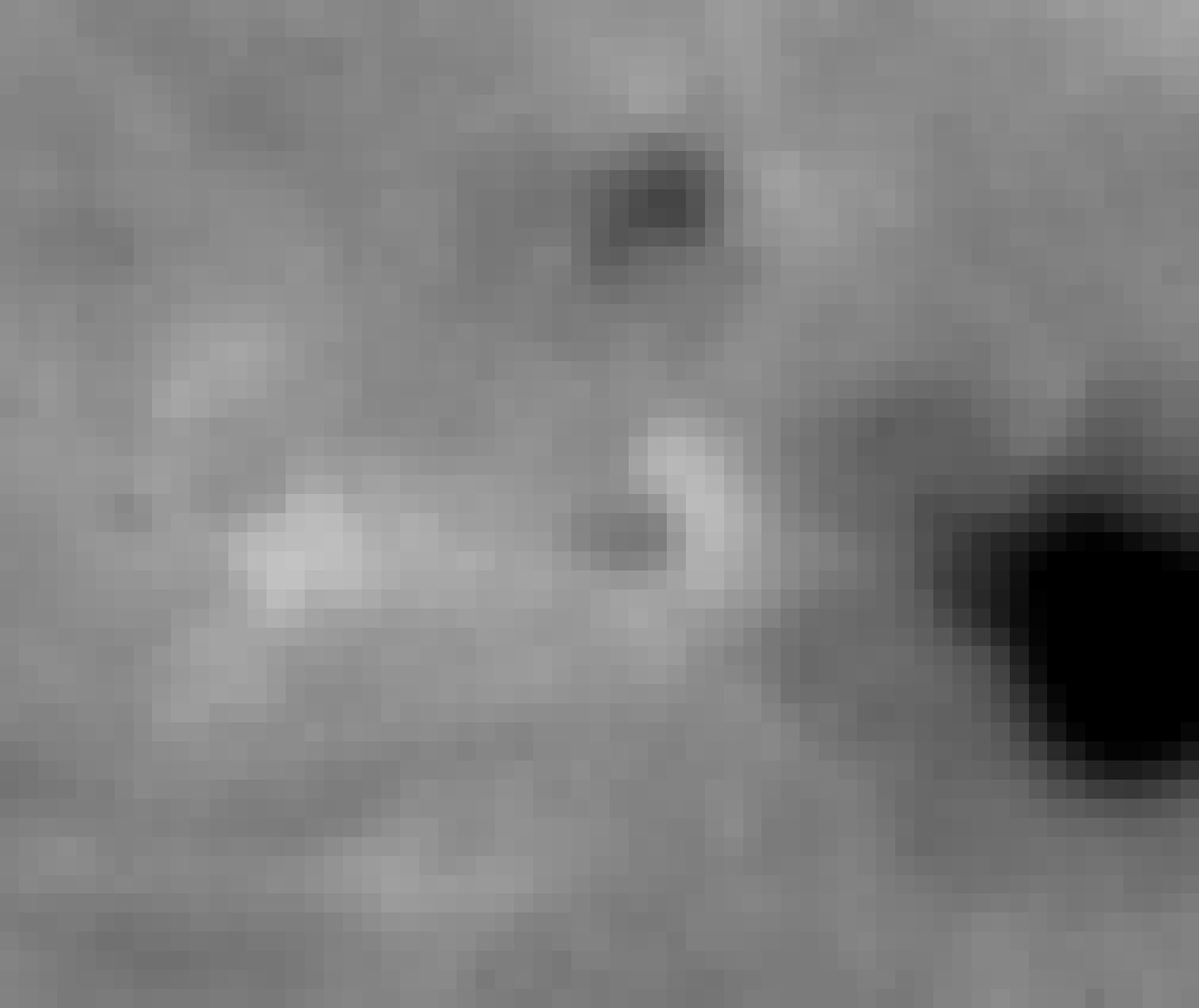} \label{subfig: LR_Final}}
    \subfloat[Final cropped GST HR image]{\includegraphics[width=0.22\textwidth]{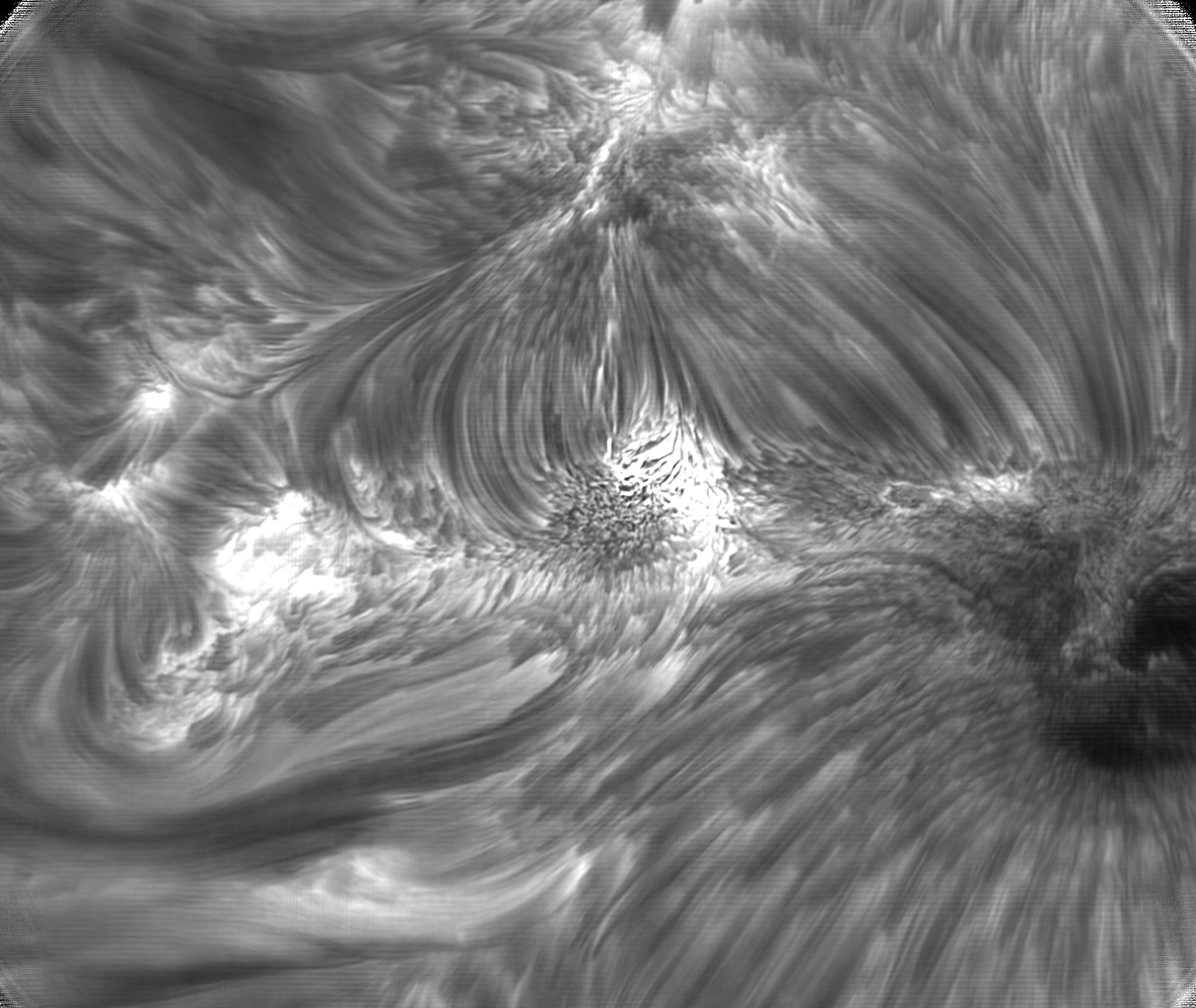} \label{subfig: HR_Final}}
    \vspace{-0.05in}
\caption{Final aligned dataset pair example: (a) GONG LR image fully prepared; (b) Corresponding GST HR image fully aligned and cropped.}  
\label{fig: LR & HR final}
\vspace{-0.25in}
\end{figure}

After rotation correction, GST HR images contained black background regions without observational data. To prevent confusing the super-resolution model during training, we cropped these non-physical regions and applied the same cropping to the corresponding GONG LR images. This ensured accurate and consistent training pairs for the super-resolution model. Figure~\ref{subfig: LR_Final} and Figure~\ref{subfig: HR_Final} show a final pair of aligned GONG LR and GST HR images after cropping.


\section{Methodology}

The methodology used in this study is based on the Enhanced Super-Resolution Generative Adversarial Network (ESRGAN)~\cite{wang2018esrgan} and Super-Resolution GAN (SRGAN)~\cite{ledig2017photo}. We employed the Real-ESRGAN model~\cite{wang2021real}, a more advanced variant of ESRGAN that is designed to produce realistic and stable HR outputs, particularly in the presence of complex or unknown image degradations. Compared to its predecessors, Real-ESRGAN incorporates a more robust network architecture, improved adversarial loss design, and perceptual loss computed from unactivated VGG features. All of these advantages can lead to sharper details and better generalization in real-world settings.

\begin{figure*}[!htbp]
\vspace{-0.1in}
  \centering
  \includegraphics[width=0.99\textwidth]{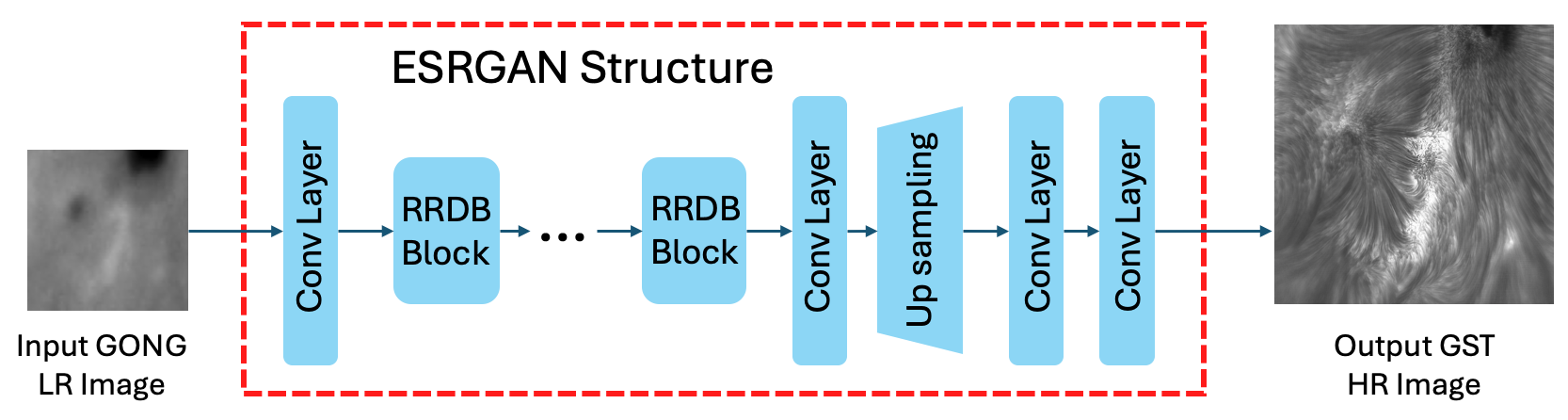} \vspace{-0.0in}
  \caption{Architecture of Real-ESRGAN used in this study. It consists of an initial convolutional layer, multiple Residual-in-Residual Dense Blocks (RRDB), followed by upsampling and additional convolutional layers to reconstruct the HR solar image.}\vspace{-0.2in}
  \label{fig:ESRGAN structure}
\end{figure*}

\subsection{Network Architecture} 

Figure~\ref{fig:ESRGAN structure} illustrates the architecture of the Real-ESRGAN generator network employed in our study. The model takes an LR GONG image as input and passes it through a convolutional layer followed by multiple Residual-in-Residual Dense Blocks (RRDBs). Each RRDB integrates both dense and residual connections, allowing the network to effectively learn rich hierarchical features. Unlike the original SRGAN framework, ESRGAN removes batch normalization layers, which helps reduce artifacts and leads to more stable training. In addition, residual scaling~\cite{lim2017enhanced} and smaller initialization weights are applied to further enhance training stability, especially for deeper networks. The extracted features are then upsampled and passed through additional convolutional layers to generate an HR image that approximates the detail level of GST observations.

\subsection{Relativistic Discriminator}

To enhance the realism of generated images, Real-ESRGAN replaces SRGAN's original discriminator with the Relativistic Average GAN (RaGAN) discriminator~\cite{jolicoeur2018relativistic}. The RaGAN discriminator evaluates image realism in a relative manner—comparing real and generated images directly rather than assessing them independently. This approach encourages the network to produce more natural textures and details. Specifically, the discriminator loss function is defined as:
\begin{equation*}
\mathcal{L}_D^{Ra} = -\mathbb{E}_{x_r}\left[\log(D^{Ra}(x_r,x_f))\right] - \mathbb{E}_{x_f}\left[\log(1 - D^{Ra}(x_f,x_r))\right],
\end{equation*}
and the corresponding generator loss is:
\vspace{-0.05in}
\begin{equation*}
\mathcal{L}_G^{Ra} = -\mathbb{E}_{x_r}\left[\log(1 - D^{Ra}(x_r,x_f))\right] - \mathbb{E}_{x_f}\left[\log(D^{Ra}(x_f,x_r))\right].
\end{equation*}
Here, $x_r$ represents real (HR) images, while $x_f$ denotes generated (fake) images from the network. The discriminator function $D^{Ra}$ calculates how realistic generated images appear relative to real images, guiding the network towards more convincing results.

\subsection{Perceptual Loss}

A significant innovation in Real-ESRGAN is its use of perceptual loss rather than relying solely on pixel-level losses. The perceptual loss measures differences between images based on features extracted from a pre-trained VGG neural network. Notably, Real-ESRGAN calculates these differences from feature maps obtained \textit{before activation layers}, resulting in richer textures, more accurate brightness, and overall improved realism. Thus, the overall generator loss combines perceptual loss, adversarial loss, and content (pixel-wise) loss ($L_1$):
\begin{equation*}
\mathcal{L}_G = \mathcal{L}_{percep} + \lambda\mathcal{L}_G^{Ra} + \eta\mathcal{L}_1,
\end{equation*}
where $\mathcal{L}_{percep}$ represents the perceptual loss from the VGG-based feature maps, $\mathcal{L}_G^{Ra}$ is the adversarial loss defined earlier, and $\mathcal{L}_1$ is the pixel-level content loss. The parameters $\lambda$ and $\eta$ balance the contributions of these three terms.

\subsection{Network Interpolation}

To achieve a practical balance between perceptual quality and reconstruction accuracy (measured typically by metrics such as Peak Signal-to-Noise Ratio, or PSNR), Real-ESRGAN uses a strategy called network interpolation. This method involves blending the parameters of two separately trained networks—one optimized for high perceptual quality (GAN-based) and the other optimized for numerical reconstruction accuracy (PSNR-based). The interpolated network parameters are computed as follows:
\begin{equation*}
\theta^{interp}_{G} = (1-\alpha)\theta^{PSNR}_{G} + \alpha\theta^{GAN}_{G},
\end{equation*}
where $\theta^{PSNR}_{G}$ are parameters of the PSNR-optimized model, and $\theta^{GAN}_{G}$ represent parameters


\section{Results}
\begin{figure}[!htbp] \vspace{-0.1in}
\centering
	\subfloat[GONG LR test image]{\includegraphics[width=0.16\textwidth]{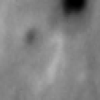} \label{subfig: LR test}}
    \subfloat[Generated HR image from GONG]{\includegraphics[width=0.16\textwidth]{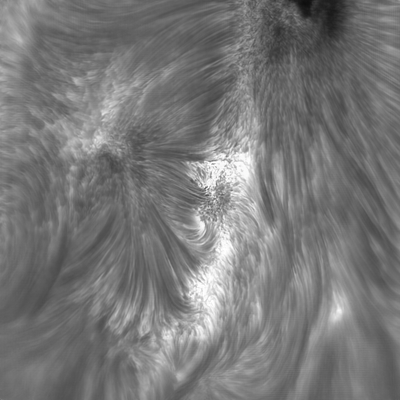} \label{subfig: HR predicted}}
    \subfloat[Ground-truth GST HR image]{\includegraphics[width=0.16\textwidth]{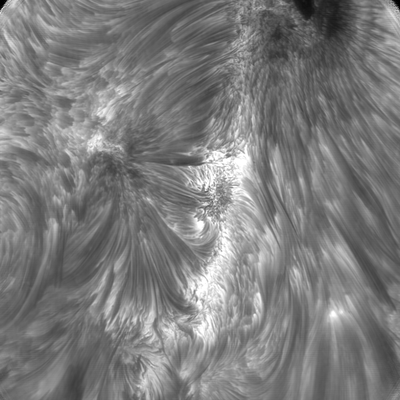} \label{subfig: HR GT}}
    \vspace{-0.05in}
    \caption{Example of the generated HR image from GONG LR image of 2023/08/31 21:35:02 (a) GONG LR testing image; (b) Generated HR image;  (c) The ground truth of the GST HR image.}  
\vspace{-0.15in}
\label{fig: SR_Result}
\end{figure}

\begin{figure}[!htbp] 
\vspace{-0.2in}
\centering
	\subfloat[Extended range GONG LR test image]{\includegraphics[width=0.23\textwidth]{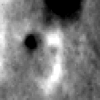} \label{subfig: LR test Extended}}
    \subfloat[Generated HR image ]{\includegraphics[width=0.23\textwidth]{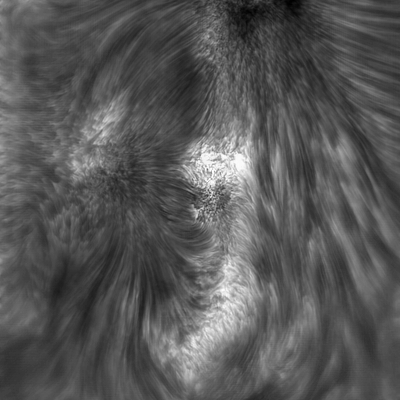} \label{subfig: HR predicted Extended}}
\vspace{-0.05in}
\caption{Example of the generated HR image from GONG LR image of extended range 2023/09/01 01:00:02 (a) GONG LR testing image; (b) Generated HR image.}  
\vspace{-0.in}
\label{fig: SR_Result_extended}
\end{figure}

Figure~\ref{subfig: LR test} shows an example of an LR GONG test image. Figure~\ref{subfig: HR predicted} presents the corresponding generated HR image obtained from our Real-ESRGAN model, while Figure~\ref{subfig: HR GT} displays the ground-truth GST observation image. 
Visually, the generated HR image demonstrates notable improvements compared to the LR input, successfully recovering intricate solar features such as sunspots and delicate filamentary structures. Despite these promising visual results, the quantitative evaluation indicates room for improvement: the average MSE is 467.15, the RMSE is 21.59, and CC is 0.7794. Although many efforts have been made to solve the alignment issues, minor misalignment still persists and negatively impacts the numerical results. In addition, we further test whether the model can generate HR outputs outside the time range covered by GST observations. Figure~\ref{subfig: LR test Extended} shows an example of an LR GONG test image for 3 hours after the end of GST observations. Figure~\ref{subfig: HR predicted Extended} shows the generated HR image obtained from our Real-ESRGAN model. Visually, it shows the model’s ability to recover structural details such as sunspot boundaries and filamentary features, which are difficult or impossible to see in the original GONG LR image Figure~\ref{subfig: LR test Extended}.

\begin{figure*}[!htbp] \vspace{-0.10in}
  \centering
\includegraphics[width=0.99\textwidth]{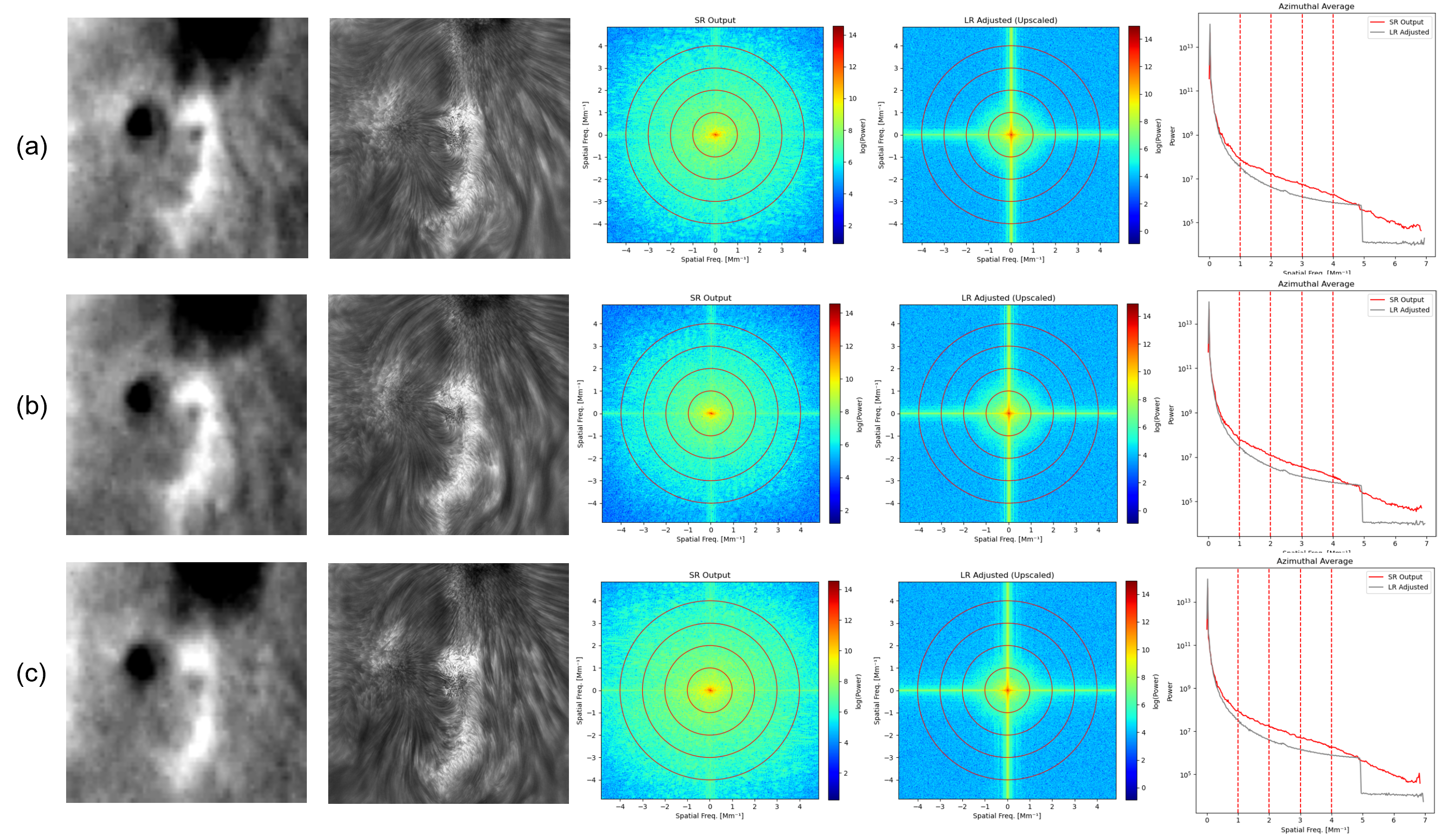} \vspace{-0in}
  \caption{Extended-range super-resolution results without ground truth. From left to right: the low-resolution input image, the corresponding SR reconstruction, the two-dimensional power spectrum of the SR output, the two-dimensional power spectrum of the upscaled LR input, and the azimuthally averaged power spectra comparing SR (red) and LR (gray). Panels (a), (b), and (c) correspond to 2023/08/31 23:00:02, 2023/09/01 00:00:02, and 2023/09/01 01:00:02, respectively.} \vspace{-0.15in}
  \label{fig: power spectra}
\end{figure*}
To further discuss the effectiveness of the extended-range SR method in recovering fine-scale solar features, we examine not only the reconstructed images but also their spatial power spectra. Visual inspection of the SR outputs already suggests clearer filaments and fibrils compared with the LR inputs, but a frequency-domain analysis is needed to confirm whether high-frequency information is truly being recovered. Figure~\ref{fig: power spectra} shows the extended-range SR results without ground truth at three consecutive time steps: Figure~\ref{fig: power spectra}(a) corresponds to 2023/08/31 23:00:02, Figure~\ref{fig: power spectra}(b) to 2023/09/01 00:00:02, and Figure~\ref{fig: power spectra}(c) to 2023/09/01 01:00:02. From left to right, each row displays the LR input image, the SR reconstructed output image, the two-dimensional power spectrum of the SR output, the two-dimensional power spectrum of the upscaled LR input, and the azimuthally averaged power spectra comparing SR (red) and LR (gray).

In these spectra, the horizontal and vertical axes represent spatial frequency, with low frequencies corresponding to large-scale background structures and high frequencies corresponding to fine details. The color scale encodes the spectral power at each frequency, with warmer colors (toward red) indicating stronger contributions and cooler colors (toward blue) indicating weaker contributions. A “better” reconstruction is expected to preserve power at low frequencies while also maintaining or enhancing power at intermediate to high frequencies, which correspond to small-scale features such as filaments and fibrils. The azimuthally averaged curves summarize this distribution: when the SR curve (red) stays above the LR curve (gray) at higher spatial frequencies, it demonstrates that the SR method successfully recovers fine-scale structure that is otherwise lost in the LR images.

\section{Conclusion and Future Work}

In this research, we successfully enhanced LR solar images from the GONG network to match the high-quality observations from the GST at BBSO. Initial visual inspections suggest that our method can effectively restore important solar details. However, current quantitative evaluations highlight alignment inaccuracies as a critical limitation. Future work will focus on further improving data alignment procedures to address this challenge. Besides, we will process and test more data rather than just for one day. Additionally, we plan to incorporate physics-informed features and explore customized enhancements to the current super-resolution model architecture, as the existing model relies primarily on previously established approaches. Lastly, we will perform local correlation tracking to determine if our method can accurately reconstruct the dynamic behavior of fine-structure features within the active region.

\section{Acknowledgment}
We gratefully acknowledge the use of data from the Goode Solar Telescope (GST) of the Big Bear Solar Observatory (BBSO). BBSO operation is supported by US NSF AGS-2309939 grant and New Jersey Institute of Technology. GST operation is partly supported by the Korea Astronomy and Space Science Institute and the Seoul National University. Data were acquired by GONG instruments operated by NISP/NSO/AURA/NSF with contributions from NOAA. This work was supported by NASA grant 80NSSC24M0174.

\bibliographystyle{IEEEtran}  %
\bibliography{references}

\end{document}